\documentclass{article}

\usepackage[final]{neurips_2025}

\usepackage{tabularx} 
\usepackage[table]{xcolor} 

\usepackage[utf8]{inputenc} 
\usepackage[T1]{fontenc}    
\usepackage{hyperref}       
\usepackage{url}            
\usepackage{booktabs}       
\usepackage{amsfonts}       
\usepackage{nicefrac}       
\usepackage{microtype}      
\usepackage{xcolor}

\usepackage{makecell}
\usepackage{multirow}
\usepackage{xspace}
\usepackage{graphicx}
\usepackage{algorithm}
\usepackage{algpseudocode}
\usepackage{booktabs}     
\usepackage{siunitx}      
\usepackage{amsmath}   
\usepackage[svgnames,table]{xcolor}
\usepackage{amsmath}

\definecolor{mylightblue}{RGB}{230,240,250}
\definecolor{tablehighlight}{RGB}{230,245,255}

\title{SVSR: A Self-Verification and Self-Rectification Paradigm for Multimodal Reasoning}

\author{%
  \textbf{Zhe Qian}\textsuperscript{1,4}, 
  \textbf{Nianbing Su}\textsuperscript{2,3},
  \textbf{Zhonghua Wang}\textsuperscript{4}, 
  \textbf{Hebei Li}\textsuperscript{5}, \\
  \textbf{Zhongxing Xu}\textsuperscript{4}, 
  \textbf{Yueying Li}\textsuperscript{6}
  \textbf{Fei Luo}\textsuperscript{7}
  \textbf{Zhuohan Ouyang}\textsuperscript{8}
  \textbf{Yanbiao Ma}{\textsuperscript\dag}\textsuperscript{7}
  \\
  \textsuperscript{1}South China Agricultural University, \textsuperscript{2}University of Glasgow,\\
  \textsuperscript{3}University of Electronic Science and Technology of China, \textsuperscript{4}Monash University,\\
  \textsuperscript{5}University of Science and Technology of China,
  \textsuperscript{6}National University of Defense Technology\\
  \textsuperscript{7}Renmin University of China\\
  \textsuperscript{8}South China Normal University\\
}

\begin{document}
\maketitle

\begin{abstract}

Current multimodal models often suffer from shallow reasoning, leading to errors caused by incomplete or inconsistent thought processes. To address this limitation, we propose Self-Verification and Self-Rectification (SVSR), a unified framework that explicitly integrates self-verification and self-rectification into the model's reasoning pipeline, substantially improving robustness and reliability in complex visual understanding and multimodal reasoning tasks. SVSR is built on a novel three-stage training paradigm. First, we construct a high-quality unified preference dataset by refining reasoning traces from pre-trained vision-language models, incorporating both forward and backward reasoning to embed self-reflective signals. Second, we perform cold-start supervised fine-tuning on this dataset to learn structured, multi-step reasoning behaviors. Third, we apply a Semi-online Direct Preference Optimization (Semi-online DPO) process, continuously augmenting the training corpus with high-quality, model-generated reasoning traces filtered by a powerful teacher VLM. This pipeline enables the model to learn, elicit, and refine its ability to self-verify and self-rectify. Extensive experiments across diverse benchmarks demonstrate that SVSR improves reasoning accuracy and enables stronger generalization to unseen tasks and question types. Notably, once trained with explicit self-reflective reasoning, the model also exhibits improved implicit reasoning ability, outperforming strong baselines even when no explicit reasoning traces are provided. These results highlight the potential of SVSR for building more dependable, introspective, and cognitively aligned multimodal systems.
\end{abstract}

\section{Introduction}
The latest advancements in Vision-Language Models (VLMs) suggest a paradigm shift from scaling up computation during training to scaling up computation at test time ~\cite{snell2024scalingllmtesttimecompute, kumar2024traininglanguagemodelsselfcorrect, qi2024mutual}. OpenAI's o1 ~\cite{o1} illustrates the effectiveness of extending test-time computation by demonstrating robust reasoning abilities through performing deep and thorough thinking, and incorporating fundamental skills such as self-checking, self-verification, self-correction, and self-exploration into the model's inference process. This paradigm not only enhances reasoning capabilities in fields like mathematics and science but also provides new insights for improving the generalization, usefulness, and safety of VLMs in various general-purpose tasks ~\cite{o1, guo2025deepseek}.

In recent years, many studies have attempted to replicate the success of o1. These efforts include using large-scale Monte Carlo Tree Search (MCTS) to construct long chain-of-thought (longCoT) training data, or scaling up test-time reasoning to improve the performance of current models ~\cite{guan2025rstar}; constructing high-quality long-CoT data for effective imitation learning; and exploring reinforcement learning to enhance the thinking abilities of VLMs on large training data and models ~\cite{guo2025deepseek, team2025kimi, cui2025process}. Recently, DeepSeek R1 ~\cite{guo2025deepseek} demonstrated that large-scale reinforcement learning can incentivize deep thinking abilities in models, with the R1 series showing the immense potential of long-thought reasoning. However, these methods typically require significant resources to enhance a model's thinking abilities, including large datasets, extensive training compute, and considerable labor and time costs. Meanwhile, for smaller or less capable VLMs, how to incentivize effective thinking, beyond distilling knowledge from more powerful models, remains unclear.

\begin{figure}[]
\centering
\includegraphics[width=0.8\textwidth]{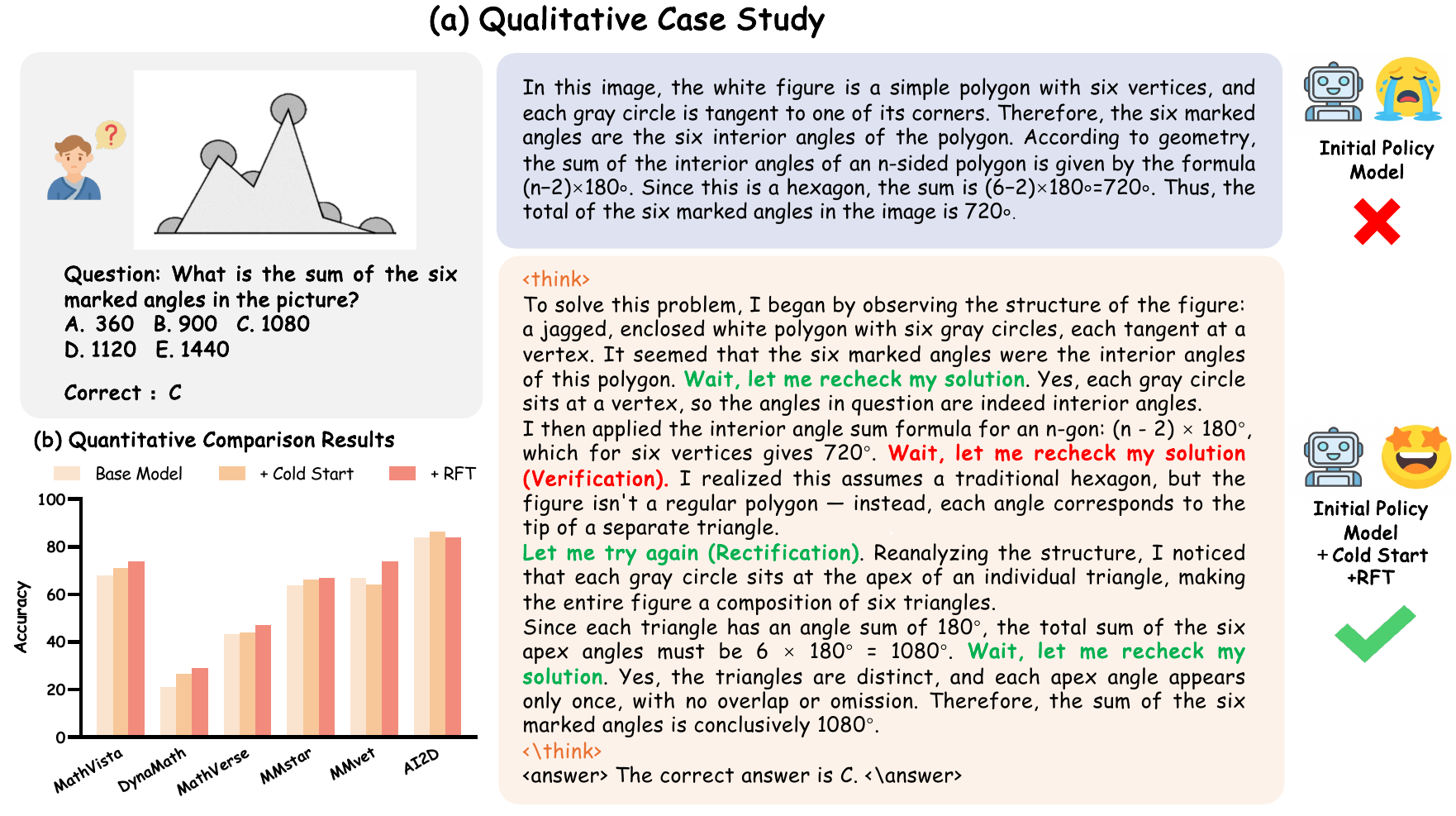}

\caption{\textbf{Impact of the Self-Verification and Self-Rectification (SVSR) framework.}
\textbf{(a) Qualitative Case Study:} Qualitative Example: Demonstrates SVSR’s ability to identify and correct an initially incorrect answer (720\textdegree) in a visual math problem, ultimately producing the correct solution (1080\textdegree) through self-verification and self-rectification..
\textbf{(b) Quantitative Comparison Results:} Reports accuracy improvements across benchmarks (e.g., MathVista, DynaMath) as the model progresses from the base model, through Cold Start SFT, to the final SVSR model with reinforcement fine-tuning.}
\label{fig:teaser}

\end{figure}

In this study, we propose SVSR, an efficient alternative designed to enhance the thinking abilities of VLMs, especially for smaller or lower-performing VLMs. Unlike having VLMs mimic the thought processes of larger, more powerful models, SVSR focuses on teaching VLMs to think deeply by iteratively employing two critical thinking skills: self-verification and self-vectification.We define correction as the combination of verification and rectification. By acquiring these two abilities, VLMs can continuously re-evaluate their solutions, identify errors during the solution exploration process, and refine previous solutions after self-checking. This paradigm also allows for flexible allocation of test-time computation to different levels of difficulty. Our results demonstrate that, using only 5k training samples, Qwen2.5-VL-7B-Instruct ~\cite{bai2025qwen25vltechnicalreport} gains significant benefits from learning self-verification and self-rectification behaviors, as supported by experimental data and comparisons with Chain-of-Thought (CoT) methods. The overall SVSR approach, detailed in Section~\ref{sec:method}, involves a three-stage training pipeline (illustrated in Figure~\ref{fig:teaser}): (1) constructing a dataset of self-correction trajectories, (2) supervised fine-tuning (SFT) for learning the reflective reasoning format, and (3) semi-online Direct Preference Optimization (semi-online DPO) for enhancing self-verification and self-rectification capabilities.

More importantly, the semi-online DPO stage in SVSR employs outcome-oriented Reinforcement Learning (RL) to further enhance these abilities. Using only a uniform preference-based dataset, RL improves the effectiveness of the self-verification and self-rectification process, enabling the model to perform more flexible and effective test-time expansion through a self-guided trial-and-error process. We used data generated by the model itself and data fused from external VLMs, and continuously iterated the dataset. By comparing fully offline and semi-online DPO, we found that semi-online DPO is not only better able to align model preferences but also ensures that the noise in the data remains within a tolerable range, thereby promoting the continuous improvement of the foundation model's reasoning abilities.

We conducted extensive experiments on Qwen2.5-VL-7B-Instruct across three mathematical reasoning benchmarks. The experiments demonstrate that SVSR outperforms competitive baselines in mathematical reasoning, including recently released example enumeration models. We also found that SVSR can be generalized to out-of-domain general tasks, such as AI2D, highlighting the effectiveness of the learned self-verification and self-rectification abilities. Furthermore, we conducted a series of analysis experiments to better showcase the reasoning mechanisms of the obtained models and provide insights for performing semi-online DPO reinforcement learning training to enhance the reasoning of large language models.

\section{Related Work}
\subsection{Self-verification and Self-rectification} 
Enabling effective self-verification and self-correction in models is a promising solution for achieving robust reasoning in Vision-Language Models (VLMs) ~\cite{madaan2024self}, and these capabilities are crucial for performing deep reasoning. Previous research suggests that directly prompting models for self-verification and self-correction is often not the optimal approach ~\cite{kumar2024traininglanguagemodelsselfcorrect}. Consequently, recent studies have explored various methods to enhance these abilities during post-training ~\cite{saunders2022self}. These methods highlight the potential of using human-annotated or model-generated data to equip models with self-verification or self-correction capabilities ~\cite{ma2025s2rteachingllmsselfverify,111}, while also indicating that behavior models alone are insufficient for effective self-verification or self-correction through supervised fine-tuning ~\cite{kumar2024traininglanguagemodelsselfcorrect,222}. In this work, we propose effective methods to enhance the self-verification and self-correction abilities of VLMs through reinforcement learning (RL) training, by fusing the model's own data with high-quality data, and we demonstrate the effectiveness of our methods through in-depth analysis.

\subsection{Reinforcement Learning for VLM Reasoning}
Reinforcement learning has proven effective in enhancing the performance of VLMs across various tasks ~\cite{liu2025visualrftvisualreinforcementfinetuning}. Recently, some studies have shown that incorporating the concept of "online" into DPO training can also effectively enhance model reasoning ~\cite{qi2024onlinedpoonlinedirect}. Recent advancements in improving models' deep thinking further leverage real-time DPO to continuously enhance the effectiveness of VLM reasoning ~\cite{qi2024onlinedpoonlinedirect, rafailov2024directpreferenceoptimizationlanguage,333}. In this work, we also demonstrate that high-quality datasets and RL frameworks can effectively improve VLM reasoning.

\subsection{Scaling Test-time Computation}
Recently, expanding test-time computation for VLM reasoning has garnered significant attention ~\cite{snell2024scalingllmtesttimecompute,444}. Existing research has explored various methods for expanding test-time computation, including: (1) aggregation-based methods, which sample multiple answers for each question and obtain the final answer by selecting the best N answers through self-consistency or using a validator or reward model ~\cite{ma2025s2rteachingllmsselfverify,555,666,777}; (2) search-based methods, which apply search algorithms such as Monte Carlo tree search, beam search, or other efficient algorithms to search for the correct trajectory; (3) iterative refinement methods, which iteratively improve test-time performance through self-improvement. In this work, we also propose an efficient framework for training VLMs to perform effective test-time expansion through iterative self-verification and self-correction. This approach can be implemented without significant effort, and the performance of self-reasoning can be continuously improved through iterative training.
\section{Method}
\label{sec:method}

\subsection{Overview}
\label{sec:overview}

\begin{figure*}[t!]
    \centering
    \includegraphics[width=0.85\textwidth]{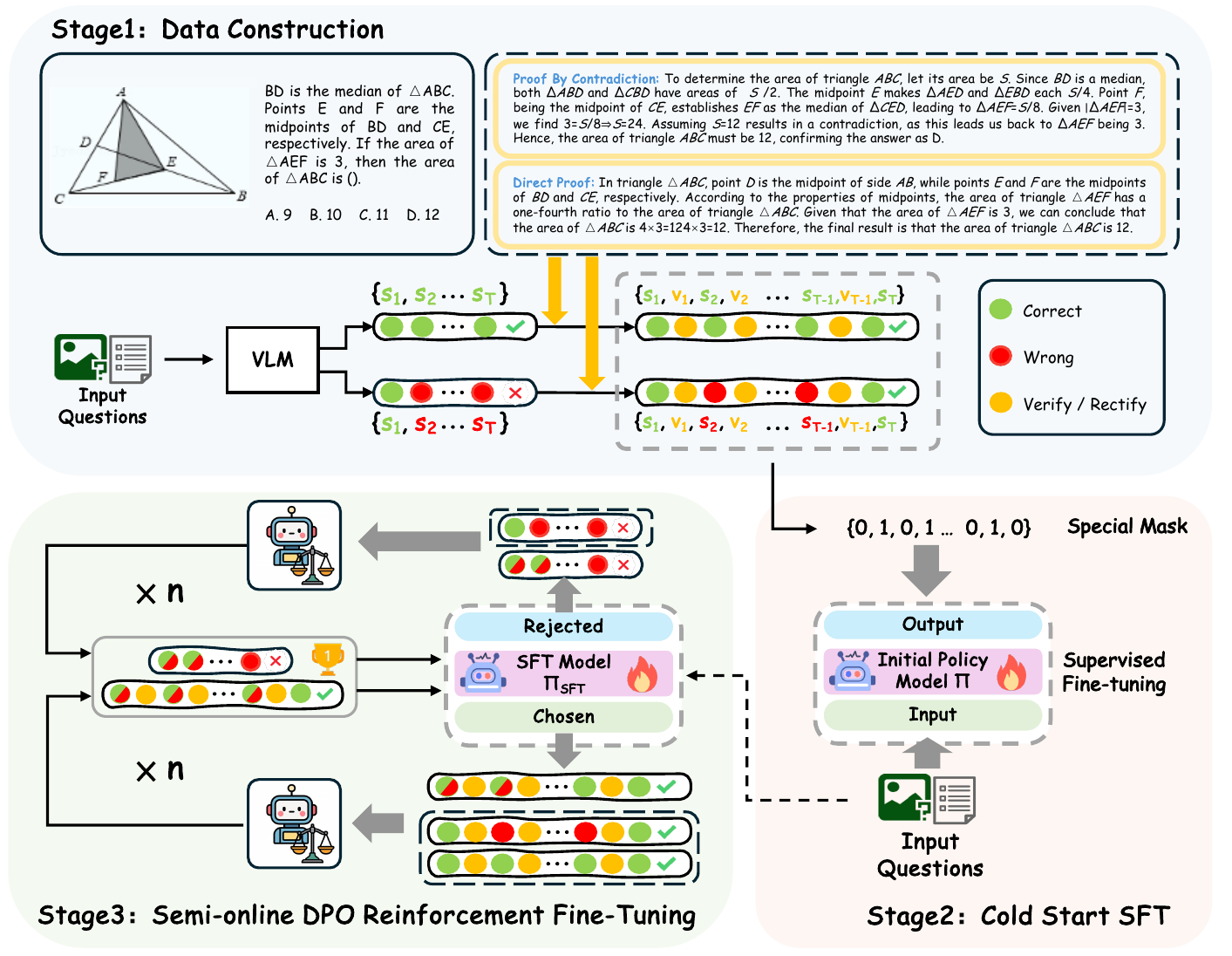}
    \caption{Overview of the SVSR three-stage training pipeline.
    \textbf{Stage 1:} Data Construction generates self-correction trajectories by refining initial reasoning through verification and rectification steps.
    \textbf{Stage 2:} Cold Start SFT fine-tunes the model on these structured trajectories, teaching it the format of reflective, multi-step reasoning.
    \textbf{Stage 3:} Semi-Online DPO further improves the model by incorporating preference data from model-generated outputs filtered by a teacher VLM, enabling iterative self-correction refinement.}
    \label{fig:svsr_framework}
\end{figure*}

In this paper, we introduce the Self-Verification and Self-Rectification (SVSR) paradigm, which equips VLMs with the ability to critically assess and revise their own reasoning steps, thereby improving the robustness and reliability of multimodal reasoning. A central challenge lies in the lack of large-scale datasets explicitly annotated for such self-reflective processes. We hypothesize that although pre-trained VLMs may have latent self-reflective capabilities, they lack both the methods to produce structured reasoning chains and the task-specific training signals required to perform self-verification and self-rectification reliably.

To address this, SVSR adopts a three-stage training pipeline, conceptually illustrated in Figure~\ref{fig:svsr_framework}.
The first stage, \textbf{Data Construction} (Stage 1), creates self-correction trajectories by processing input questions through an initial policy. Correct reasoning paths are retained (visualized in green), while incorrect ones (red) undergo iterative self-verification and self-rectification, represented by yellow "Verify / Rectify" nodes. This results in structured sequences like ${S_1, V_1, S_2, V_2, \dots, S_T}$. The figure also includes an example of a visual math problem, showcasing two distinct reasoning paths - direct reasoning and contradiction-based reasoning - to more clearly illustrate the problem-solving process.
In \textbf{Cold Start SFT} (Stage 2), the model is fine-tuned on these structured trajectories using supervised learning. A binary mask (e.g., ${0, 1, 0, 1, \dots}$) distinguishes between reasoning and verification steps, guiding the model to learn the self-reflective reasoning format.
In Stage 3, \textbf{Semi-Online DPO Fine-Tuning}, the fine-tuned model $\pi_{\text{SFT}}$ generates multiple reasoning trajectories (denoted "$\times n$") per input. A teacher VLM ranks these outputs, from which high-quality "chosen" "rejected" pairs are selected. These dynamic preference samples are incorporated into the training loop to iteratively refine the model's judgment and self-correction ability.

\subsection{Stage 1: Self-Correction Trajectory Dataset Construction}
\label{sec:data_construction}

The first stage, \textbf{Self-Correction Trajectory Dataset Construction}, builds $\mathcal{D}_{\text{SFT}}$ to train VLMs for explicit self-verification and self-rectification, as introduced in Section~\ref{sec:overview}. Starting with a \textit{unified preference dataset} from sources like VQAv2, ScienceQA, we generate \textit{self-correction trajectories} using direct derivation and proof-by-contradiction, reflecting Stage 1 in Figure~\ref{fig:svsr_framework}. Each trajectory $Y$ for an input $(I, Q)$ is $Y = (s_1, v_1, \dots, s_k, v_k, A_{\text{final}})$, where $s_i$ is a reasoning step, $v_i$ is its verification/rectification (cf. $S_i, V_i$ in Figure~\ref{fig:svsr_framework}), $k$ is the correction cycle count ($k=0$ if initially correct), and $A_{\text{final}}$ matches the ground truth $A_{\text{GT}}$.

To create $\mathcal{D}_{\text{SFT}}$, a generator VLM ($M_{\text{gen}}$, e.g., GPT-4o) processes each instance $(I, Q, A_{\text{GT}})$. If $M_{\text{gen}}$'s initial reasoning is incorrect, or to create explicit reflective examples (Figure~\ref{fig:svsr_framework}), it is iteratively prompted to review its prior step $s_i$, identify errors, and generate corrective reasoning $v_i$. This process forms $(s_i, v_i)$ pairs until $A_{\text{GT}}$ is reached or a maximum step limit, populating $\mathcal{D}_{\text{SFT}}$ with diverse examples of direct and corrected reasoning.

To ensure dataset quality for behavior initialization, trajectory generation adheres to key principles, centrally a rule-based state transition. This dictates that a \textbf{"solve"} step is followed by \textbf{"verify"}; an \textbf{"INCORRECT"} verification triggers \textbf{"rectify"}; and a \textbf{"CORRECT"} verification or rectification leads back to \textbf{"solve"}. Trajectories conclude, for instance, when a verification step's output indicates a rectification is finalized. Further principles include promoting reasoning diversity by varying correction cycle lengths ($k \in \{1,2,3,4\}$), filtering incoherent or unsuccessful attempts, and aligning trajectory length with problem difficulty (based on $M_{\text{gen}}$'s initial accuracy). This process yields $\mathcal{D}_{\text{SFT}}$, a dataset of $(I, Q, Y)$ tuples encoding valid, diverse self-correction processes that reflect both problem difficulty and structured reasoning.

{\small\[
\label{eq:state_func}
    Type(a_{i+1}) = \begin{cases}
    \text{verify}, & Type(a_T) = \text{solve} \\
    \text{rectify}, & Type(a_T) = \text{verify} \text{ and } Parser(a_T) = \text{INCORRECT} \\
    \text{solve}, & Type(a_i) = \text{verify / rectify} \text{ and } Parser(a_i) = \text{CORRECT} \\
    \text{<end>}, & Type(a_i) = \text{verify and answer's verb is rectified}
\end{cases}
\]}

\subsection{Stage 2: Cold Start SFT for Reasoning Format Initialization}
\label{sec:cold_start}
While VLMs may possess latent reflective capabilities, they often lack the structured output format required to explicitly express step-by-step self-verification and self-rectification. The Cold Start SFT stage addresses this gap by initializing a base VLM through supervised training on the structured dataset $\mathcal{D}_{\text{SFT}}$ constructed in the previous stage.

We initialize the base model $\pi_{\text{base}}$ using a pre-trained VLM architecture. The training objective is to maximize the likelihood of generating the target self-correction trajectory $Y$ conditioned on the input image $I$ and question $Q$. This is achieved using the standard autoregressive cross-entropy loss:
\[
\label{eq:sft_loss}
\mathcal{L}_{\text{SFT}}(\theta) = - \sum{(I, Q, Y) \in \mathcal{D}_{\text{SFT}}} \log P_{\theta}(Y\mid I, Q),
\]
where $\theta$ denotes the parameters of the VLM $\pi_{\theta}$. This phase primarily enables the model to internalize the syntax, structure, and flow of self-corrective reasoning chains. Specifically, it learns to generate sequences $Y$ where tokens $a_t$ corresponding to different action types (such as `solve', `verify', and `rectify', as defined by the trajectory generation process in Section~\ref{sec:data_construction}) are produced in the correct order and format. The model thus learns to explicitly articulate each stage of the reflective reasoning process. The resulting model, trained to follow this desired output format, is denoted as $\pi_{\text{SFT}}$.

\[
\label{eq:mask}
\delta_{mask}(a_t) = \begin{cases}
    1, & \text{if } Type(a_t) = \text{verify / rectify} \\
    0, & \text{otherwise}
\end{cases}
\]

\subsection{Stage 3: Semi-Online DPO for Capability Enhancement}

\label{sec:semi_online_dpo}
Although the SFT stage effectively teaches the model the desired self-corrective reasoning format, it may fall short in instilling a strong preference for genuinely correct and well-justified reasoning paths over superficially plausible but ultimately flawed alternatives. Moreover, the static nature of $\mathcal{D}_{\text{SFT}}$ limits its ability to adapt to the model's evolving preferences during training, preventing it from capturing the dynamic nuances needed for robust generalization across diverse scenarios. To address these limitations and further refine the model's judgment, we adopt DPO within a semi-online training framework.

\subsubsection{Background of DPO}
\label{sec:dpo_background}
DPO provides an efficient alternative to traditional Reinforcement Learning from Human Feedback (RLHF) by directly optimizing the policy model $\pi_\theta$ using preference data, thereby eliminating the need to train a separate reward model. Given a dataset of pairwise preferences $\mathcal{D}_{\text{pref}} = \{(I, Q, Y_{\text{win}}, Y_{\text{lose}})\}$, where $Y_{\text{win}}$ and $Y_{\text{lose}}$ denotes the preferred and dispreferred reasoning trajectory for the same input $(I, Q)$, DPO aims to minimize the following loss:
{\small\[
\label{eq:dpo_loss}
\mathcal{L}_{\text{DPO}}(\theta; \pi_{\text{ref}}) = - \mathbb{E}_{(I, Q, Y_{\text{win}}, Y_{\text{lose}}) \sim \mathcal{D}_{\text{pref}}} \left[ \log \sigma \left( \beta \log \frac{\pi_{\theta}(Y_{\text{win}} | I, Q)}{\pi_{\text{ref}}(Y_{\text{win}} | I, Q)} - \beta \log \frac{\pi_{\theta}(Y_{\text{lose}} | I, Q)}{\pi_{\text{ref}}(Y_{\text{lose}} | I, Q)} \right) \right],
\]}
where $\pi_{\text{ref}}$ is a fixed reference model, typically initialized as the SFT model $\pi_{\text{SFT}}$. The function $\sigma$ denotes the sigmoid function, and $\beta$ is a hyperparameter that modulates the divergence from the reference policy.

\subsubsection{Semi-Online Data Generation and Preference Labeling}
\label{sec:semi_online_data}
While conventional DPO (Section~\ref{sec:dpo_background}) relies on a static preference dataset $\mathcal{D}_{\text{pref}}$, this can lead to data staleness as the policy $\pi_{\theta}$ evolves, diminishing the dataset's relevance to the model's current weaknesses. Our Semi-online DPO approach addresses this by periodically refreshing the training data. At scheduled intervals, we use the current policy $\pi_{\theta}$ to generate $N$ diverse candidate trajectories $\{Y_1, \dots, Y_N\}$ for prompts $(I, Q)$ sampled from $\mathcal{D}_{\text{source}}$. These candidates are then evaluated by a powerful teacher VLM ($M_{\text{teacher}}$) based on criteria including final answer correctness, logical coherence, and the quality of self-verification within the trajectory (encompassing accuracy of verification claims and effectiveness of verification steps). The teacher provides preference judgments, enabling the selection of a preferred trajectory $Y_{\text{win}}$ and a dispreferred one $Y_{\text{lose}}$. After filtering out low-quality or ambiguous pairs, this process yields a dynamic, high-quality preference dataset $\mathcal{D}_{\text{online}}$ that continually reflects the model's current capabilities and failure modes, thereby ensuring more effective DPO updates.

\subsubsection{Iterative Training Mechanism}
\label{sec:iterative_training}

The Semi-online DPO training follows an iterative refinement loop centered around a dynamic buffer $\mathcal{B}_{\text{pref}}$ that stores preference data for optimization. This buffer is initially seeded with preference pairs derived from the SFT dataset $\mathcal{D}_{\text{SFT}}$, where correct trajectories are designated as $Y_{\text{win}}$ and paired with either synthetically generated incorrect variants or lower-quality trajectories identified during Stage 1 as $Y_{\text{lose}}$. A small set of manually curated high-quality pairs may also be included to bootstrap training. 

The training proceeds over $T$ iterations. In each iteration $t$, the process consists of three steps:

\textbf{Online Preference Generation:}
We use the current policy model $\pi_{\theta_t}$ and the teacher VLM $M_{\text{teacher}}$ (as described in Section~\ref{sec:semi_online_data}) to generate a batch of new preference pairs $\mathcal{D}_{\text{online}}$.

\textbf{Buffer Update:}
The preference buffer $\mathcal{B}_{\text{pref}}$ is updated by adding the newly generated preference pairs and optionally removing outdated or low-utility pairs to maintain a relevant dataset within a maximum size $B_{\text{max}}$. This can be implemented via FIFO replacement or adaptive strategies based on training loss contribution or preference score differentials.

\textbf{Policy Optimization:}
For a fixed number of training steps $S$, we sample mini-batches $\mathcal{B}_{\text{batch}}$ from $\mathcal{B}_{\text{pref}}$ and update the model parameters $\theta_t$ via gradient descent on $\mathcal{L}_{\text{DPO}}(\theta_t; \pi_{\text{ref}})$, where the reference model $\pi_{\text{ref}}$ is fixed as $\pi_{\text{SFT}}$. After $S$ steps, we obtain an updated model $\pi_{\theta_{t+1}}$, and the cycle repeats.

This iterative refinement allows the policy $\pi_\theta$ to continuously adapt to its evolving capabilities, leveraging dynamic preference signals to improve its self-verification and self-rectification behaviors. The effectiveness of this process depends on several hyperparameters, including the DPO strength coefficient $\beta$, learning rate, batch size, data regeneration frequency (e.g., every $M$ steps or per buffer epoch), and the buffer management strategy.  The final output of this stage is the optimized policy model $\pi_{\theta_T}$.
\section{Experiments}\label{exp}
\label{experiments}
\subsection{Experimental Setup}
To assess the effectiveness of our method, we conducted experiments using Qwen2.5-VL-7B-Instruct ~\cite{bai2025qwen25vltechnicalreport}. We extensively fine-tuned the model not only on complex mathematical reasoning tasks but also across a broad spectrum of multimodal reasoning challenges—including visual question answering, diagram interpretation, and textual entailment involving images—in order to comprehensively assess the robustness, adaptability, and generalizability of our proposed approach across varied reasoning domains.

\textbf{Datasets }  
We constructed dedicated datasets for the two fine-tuning stages. \textbf{Phase 1: Cold start.} We leveraged a collection of existing datasets as references, including MathVision (3K, ~\cite{MathVisionDataset2024}), MMR1 (7K, ~\cite{MathMR1Dataset}), WaltonFuture (12K, ~\cite{WaltonFutureMath12kImage}), Ayush-Singh/maths-vision-task-splits (800, ~\cite{SinghMathsVisionTaskSplits800}), We-Math (1.7K, ~\cite{WeMathDataset}), and CoMT/creation (500, ~\cite{CoMTCreation500}). Based on these datasets, we employed GPT-4o ~\cite{hurst2024gpt} to construct an initial unified preference dataset, which captured fundamental reasoning preferences. Subsequently, GPT-4o~\cite{GPT4oUnifiedPreference2024} was used to generate both positive and negative self-verification signals for each sample. These were combined with the initial preferences to form a high-quality, verification-enhanced preference dataset. We selected 5K high-quality samples from this dataset for the cold start phase, while the remaining examples were reserved for subsequent reinforcement fine-tuning. \textbf{Phase 2: Reinforcement fine-tuning.} Building on the dataset constructed in Phase 1, we further augmented the training corpus with refined datasets, including DPO-ScienceQA (20K, ~\cite{LongpreDPOScienceQA2023}) and optionally nano-omni-vlm-dpo (200K, ~\cite{NanoOmniVLMdpo}). In addition, we leveraged the model itself to generate high-quality unified preference data from selected samples, and employed a teacher VLM to score these generations. We retained only the top-rated rejected and chosen pairs to ensure data quality. To enhance the efficiency and stability of reinforcement fine-tuning, we filtered out overly simple or excessively difficult instances. This process resulted in a curated RL training set of approximately 20K high-quality examples. 

\textbf{Baseline }  
We compare the proposed method with strong baselines from the following three categories:
\textbf{1) State-of-the-art VLMs.} This includes leading commercial models such as GPT-4o, GPT-4V, Claude-3.5 Sonnet, Claude-3.7 Sonnet, and Google's Gemini 2.0 series (Flash and Pro).
\textbf{2) Top-performing open-source reasoning models.} We evaluate against high-performing open-source models with strong reasoning capabilities, including InternVL2.5-8B ~\cite{chen2025expandingperformanceboundariesopensource}, InternVL2.5-26B ~\cite{chen2025expandingperformanceboundariesopensource}, InternVL3-9B ~\cite{zhu2025internvl3exploringadvancedtraining}, Qwen2-VL-7B ~\cite{wang2024qwen2vlenhancingvisionlanguagemodels}, MiniCPM-V2.6 ~\cite{hu2024minicpmunveilingpotentialsmall}, Cambrian-34B ~\cite{tong2024cambrian1fullyopenvisioncentric}, and LLaVA-OneVision-72B ~\cite{li2024llavaonevisioneasyvisualtask}.
\textbf{3) SFT with CoT variants.} To examine the effectiveness of our self-rectification paradigm, we compare against models trained using various CoT constructions. For a fair comparison with our model, the training data size is matched to our cold-start dataset.

\textbf{Evaluation Benchmarks }
To comprehensively evaluate the effectiveness of our method, we selected benchmarks across three domains. 
1) For mathematical reasoning, we adopted three widely used datasets: MathVista ~\cite{LuMathVista2023}, DynaMath ~\cite{DynaMathDataset}, and MathVerse ~\cite{MathVerseDataset}. 
2) To assess performance on chart and document understanding, we used AI2D ~\cite{AI2DDataset2016}, which focuses on diagram interpretation and visual structure comprehension. 
3) For evaluating general multimodal understanding and robustness to hallucinations, we employed MMStar ~\cite{MMStarDataset} and MMVet ~\cite{MMVetDataset}. 
These benchmarks collectively cover a diverse range of reasoning modalities, from structured visual input to open-ended multimodal queries. Detailed descriptions of each dataset are provided in the Appendix.

\textbf{Evaluation Metrics }
We report Pass@1 accuracy as the primary evaluation metric across all baselines. For inference, we utilize vLLM and implement our evaluation pipeline based on the Qwen2.5-VL ~\cite{bai2025qwen25vltechnicalreport} codebase. Full implementation details, including hyperparameter configurations and infrastructure settings, are provided in the Appendix.

\subsection{Quantitative Comparisons with Baselines}

\begin{table}[htbp]
\setlength{\tabcolsep}{3pt}
    \centering 
    \caption{Model performance on multimodal benchmarks. Scores represent Pass@1 accuracy (\%). The best result in each benchmark column is shown in \textbf{bold}.}
    \label{model_benchmark_table}
    \small
    \sisetup{
        detect-weight=true, 
        mode=text,          
        table-format=2.1    
    }
    \begin{tabularx}{\linewidth}{@{}X*{7}{S}@{}}
        \toprule
        Model & {MathVista} & {DynaMath} & {MathVerse} & {MMstar} & {MMvet} & {AI2D} & {Average} \\
        \midrule
        \multicolumn{8}{c}{\textbf{Closed-Source Models}} \\
        \midrule 
        GPT-4V                   & {--} & {--} & {--} & 56.0 & 67.5 & 78.2 & {--} \\
        GPT-4o-20240513          & 60.0 & 34.5 & 40.6 & 64.7 & 69.1 & 84.6 & 59.0 \\
        Claude-3.5-Sonnet               & 64.7 & 36.6 & 44.2 & 65.1 & 70.1 & 81.2 & 60.3 \\
        Gemini-1.5-Pro             & 68.7 & \textbf{40.7} & 30.1 & 59.1 & 64.0 & 79.1 & 57.0 \\
        \midrule
        \multicolumn{8}{c}{\textbf{Open-Source Models}} \\
        \midrule 
        InternVL2.5-8B \cite{chen2025expandingperformanceboundariesopensource}         & 64.5 & 9.4 & 22.8 & 62.8 & 62.8 & 84.5 & 51.1 \\
        MiniCPM-V2.6 \cite{hu2024minicpmunveilingpotentialsmall}                  & 60.8 & 9.8 & 18.9 & 57.5 & 60.0 & 82.1 & 48.2 \\
        InternVL3-9B \cite{zhu2025internvl3exploringadvancedtraining}                  & 71.5 & 26.7 & 35.3 & 66.3 & 76.2 & 84.6 & 60.1 \\
        Qwen2-VL-7B \cite{wang2024qwen2vlenhancingvisionlanguagemodels}            & 62.3 & 18.1 & 30.1 & 60.7 & 62.0 & 83.0 & 52.7 \\
        Cambrian-34B \cite{tong2024cambrian1fullyopenvisioncentric}              & 53.2 & {--} & {--} & 54.2 & 53.2 & 79.5 & {--} \\
        LLaVA-OneVision-72B \cite{li2024llavaonevisioneasyvisualtask}      & 67.1 & 15.6 & 27.2 & 65.8 & 60.6 & 85.6 & 53.7 \\
        InternVL2.5-26B \cite{chen2025expandingperformanceboundariesopensource}                & 68.2 & 11.4 & 24.0 & 66.5 & 65.0 & 86.4 & 53.6 \\
        \midrule
        \multicolumn{8}{c}{\textbf{Base Model}} \\
        \midrule 
        Qwen2.5-VL-7B-Instruct \cite{bai2025qwen25vltechnicalreport}& 68.1 & 21.0 & 43.3 & 63.9 & 67.1 & 83.9 & 57.9 \\
        base (w/ CoT)& 68.2 & 23.7 & 43.9 & 61.8 & 67.2 & 83.1 & 58.0 \\
        \midrule
        \multicolumn{8}{c}{\textbf{Ours}} \\
        \midrule 
        \rowcolor{tablehighlight} Qwen2.5-VL-7B-Instruct& 68.1 & 21.0 & 43.3 & 63.9 & 67.1 & 83.9 & 57.9 \\
        \rowcolor{tablehighlight} +cold start      & 71.1 & 26.8 & 44.1 & 66.2 & 64.2 & \textbf{86.5} & {59.8} \\
        \rowcolor{tablehighlight} +RFT      & {\bf 73.8} & 29.2 & {\bf 47.2} & {\bf 67.1} & {\bf 73.8} & 84.1 & \textbf{62.5} \\        
        \bottomrule
    \end{tabularx}
\end{table}

We present the main resuls in Table~\ref{model_benchmark_table}, comparing our SVSR framework against baselines. Notably, incorporating self-verification and self-rectification significantly improves performance on most tasks compared to the base model, Qwen2.5-VL-7B-Instruct ~\cite{bai2025qwen25vltechnicalreport}, with particularly strong gains in reasoning-intensive benchmarks. This aligns with the core intuition that self-corrective CoT reasoning fosters a deeper understanding of visual content. Although the model is primarily trained on reasoning tasks, we observe that the learned self-verification and self-rectification abilities generalize well to out-of-distribution domains, as shown in the table. After SVSR training, the model demonstrates improvements over the base model on most tasks and achieves performance comparable to or even exceeding that of several strong baselines. These findings suggest that self-verification and self-rectification are not merely task-specific heuristics, but rather transferable cognitive abilities that enhance general reasoning. Moreover, we expect that applying SVSR training to data from other domains may lead to further performance gains in specialized settings. 

\subsection{Exploring Self-Verification and Self-Rectification Capabilities}

\subsubsection{Bias and Effectiveness in Verification Approaches}
We summarize two key findings from our experiments comparing direct derivation and proof by contradiction as verification strategies. (1) Generally, direct derivation achieved higher overall verification accuracy than proof by contradiction. This outcome is intuitive, as most existing models are trained for forward problem-solving. Simultaneously, we also observed in the dataset that, regardless of the prompt, the model consistently tends to use direct inference to solve problems.  (2) While accuracy provides a basic measure, it is insufficient to fully capture the effectiveness of verification. For instance, when answer accuracy is already high, labeling all predictions as correct direct proof verification scores without offering meaningful insight. Upon further analysis, we observed that problem-solving-based verification exhibited a pronounced bias toward confirming answers as correct, whereas contradiction-based verification produced more balanced outcomes. We attribute this bias to the influence of prior reasoning steps in direct derivation, consistent with prior findings that LLMs struggle to detect their own mistakes. In contrast, proof by contradiction approaches verification from a different angle, mitigating the influence of prior solutions. To minimize methodological bias, we randomly alternated between the two strategies during behavior initialization.

\subsubsection{Quantifying Self-Verification and Self-Rectification Behaviors}

To evaluate the effect of SVSR training on the model's self-verification and self-rectification abilities, we introduce specific behavioral metrics, with results presented in Table~\ref{name}.

\textbf{For self-verification,} we report two measures: (1) Verification Accuracy, which measures the overall accuracy of the model when performing self-verification. Specifically, it refers to the overall accuracy rate of the model in judging whether its own reasoning steps are correct. If the model can accurately identify correct and incorrect steps, the verification accuracy will be high. (2) Error Recall, which measures the proportion of times a model successfully identifies an error when its previous answer (or reasoning step) was incorrect. In other words, error recall measures how likely a model is to detect a mistake after it has already made one. The higher this metric, the better the model is at identifying errors in its reasoning process. Table~\ref{name} illustrates Verification Accuracy and Error Recall. The SFT+RFT approach shows notable improvements in these aspects compared to SFT alone. 

\textbf{For self-rectification,} we define (1) Error-to-Correct Ratio, representing how often the model successfully revises an initially wrong answer into a correct one, and (2) Correct-to-Error Ratio, indicating the frequency of unnecessary revisions that degrade a correct answer. As shown in Table~\ref{name} (right), the SFT+RFT model exhibits an enhanced capability in the "Incorrect to Correct" metric and a desirable reduction in the "Correct to Incorrect" metric when compared to the SFT baseline. These metrics jointly characterize the model's capacity to reflect on its own reasoning and make beneficial or detrimental changes during inference, with SFT+RFT showing superior performance in these self-correction behaviors.


\begin{table}[]
\centering
\small
\caption{Quantitative results of self-verification and self-rectification abilities across model variants.}
\label{name}
\tabcolsep=0.1cm{
\begin{tabular}{@{}lcccc@{}}
\toprule
\multirow{2}{*}{Model}     & \multicolumn{2}{c}{Self-verification} & \multicolumn{2}{c}{Self-rectification}      \\ \cmidrule(l){2-5} 
                           & Verification Acc  & Error Recall & Incorrect to Correct & Correct to Incorrect \\ \midrule
Qwen2.5 VL-7B-Instruct & 39.87             & 48.18             & 10.15                & 1.22                 \\
+ Cold Start               & 56.12             & 61.92             & 16.81                & 5.31                 \\
+ RFT                      & 69.24             & 69.15             & 23.10                & 2.89                 \\ \bottomrule
\end{tabular}
}

\end{table}

\subsubsection{Adaptive Inference Behavior Across Difficulty Levels}

\begin{figure}[t]
    \centering
    \includegraphics[width=0.75\linewidth]{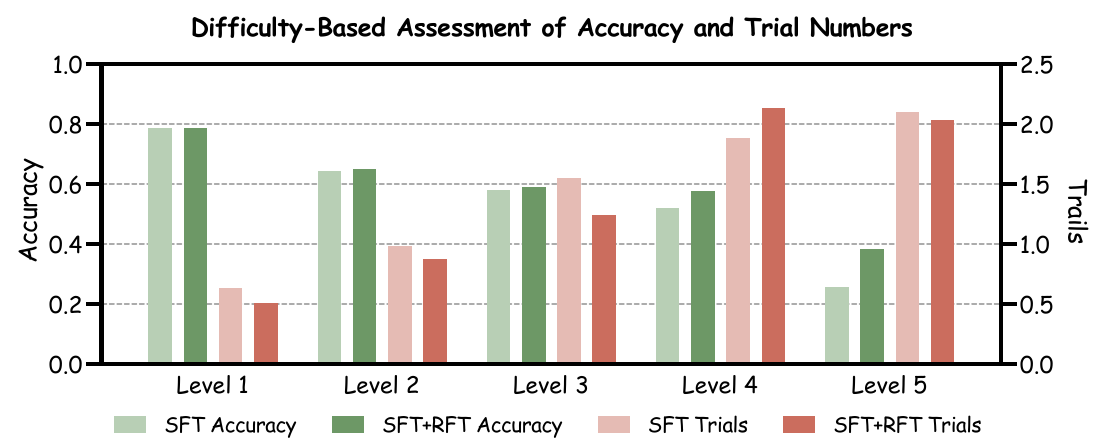}
    \caption{Difficulty-based assessment of model accuracy and average number of trials. The bars (left axis) show the accuracy of models across five difficulty levels, while the lines (right axis) represent the average number of verification attempts required. The SFT+RFT model demonstrates both higher accuracy and more adaptive resource allocation, using fewer trials on easier problems (Levels 1-2) while investing more verification steps on difficult problems (Levels 3-5).}
    \label{fig:diff_acc_trials_nums_figure}
\end{figure}

To further illustrate the effect of SVSR training, we analyze the model's behavior across different problem difficulty levels. As shown in Figure~\ref{fig:diff_acc_trials_nums_figure}, we report two metrics for each level: answer accuracy and the average number of trials, where each trial refers to a full correction loop in which the model identifies an incorrect answer via self-verification, attempts to revise it, and then re-validates the result. If the rectification fails, the process repeats, initiating another trial.
Two key observations emerge:
(1) Answer accuracy on harder problems—particularly Level 5—improves significantly after reinforcement fine-tuning, indicating that SVSR enhances the model's reasoning robustness under more challenging conditions.
(2) After reinforcement training, the model improves its correction accuracy and reduces the number of trials for easier problems (Levels 1–3). For more difficult problems, however, the number of trials remains high or even increases slightly, suggesting that the model prioritizes finding correct answers over minimizing effort when uncertainty is high.

\section{Conclusion}
This study introduces SVSR, a pioneering visual language model that integrates self-verifying and self-rectifying reasoning, aiming to significantly enhance visual understanding and self-reflection capabilities. To achieve this, we implement an innovative exploratory trial-and-error reinforcement fine-tuning method. This approach is specifically designed to elicit and incentivize the model's inherent capacity for self-verification, allowing it to learn and improve through iterative feedback. The methodology comprises three key stages: data construction, cold start, and semi-online DPO (Direct Preference Optimization) reinforcement learning.

Despite progress, the accuracy of the model's self-verification and self-rectification mechanisms in complex tasks remains a bottleneck. Future research will focus on addressing this issue to optimize model performance and robustness. A detailed discussion can be found in the appendix.



\bibliographystyle{unsrt}
\bibliography{reference.bib}

\newpage
\appendix
\section{Further Methodological Insights}
\textit{Building the Dataset.}
Given the sampling from cutting-edge VLMs, we initially constructed a unified preference dataset using CoT-prompted VLMs. To build more effective validation and correction, we informed the VLM of the correct answer in advance and forced it to answer in our prescribed format. At the same time, we provided the VLM with two basic methods for verifying its thinking, namely, positive and negative arguments. We found that, despite being instructed to output answers in the prescribed self-verification and self-rectification format, most still lacked the subconsciousness of self-verification and self-rectification, and could not well integrate reflective thinking into their thought processes. Finally, we collected a complete unified preference dataset of self-verification and self-rectification by querying GPT-4o, which demonstrated a strong ability to follow instructions, could answer using the prescribed format, and could adopt reasonable verification methods such as forward and reverse thinking and inductive reasoning. For these collected prompts, we performed a brief filtering, discarding any data that did not conform to the self-verification and self-rectification format. By refining and filtering the data, we constructed SFT data as follows. For each question's "chosen" answer, we determined the number of answering attempts required to finally obtain the correct answer based on the accuracy of the initial sampling. The lower the accuracy, the more generated answer arguments. In our implementation, we divided all questions into 5 difficulty levels and constructed multi-round verification and rectification answer sequences based on decreasing accuracy. At the end of each step, we appended "Wait, let me recheck my solution" followed by the answer verification of the previous step. If an error was found in each verification, we appended "Let me try again" followed by the rectification. In the "chosen" construction part of the dataset, we ensured that the last answer in the sequence was correct.

\textit{Cold-starting with only the "chosen" data from the unified preference dataset.}
Why we need it. In the cold-start phase, we use a small amount of high-quality, self-verifying, and self-correcting image generation reasoning data to allow the model to internalize the format and structure of CoT (Chain-of-Thought) reasoning. With cross-task prior knowledge, cold-starting can generate accurate self-verification and self-correction-based reasoning processes for many simple scenarios. In addition, we have the model focus on each twist and turn of self-verification and self-rectification. The model explores different reasoning paths within this format and gradually learns to prefer more multi-dimensional solutions.

\textit{Semi-Online DPO Fine-tuning of the Model.}
After cold-starting, the model has internalized the CoT reasoning format; however, this is not sufficient, and these datasets cannot align the model's preferences well. To address this, we re-established a new dataset. This dataset is divided into multiple sub-datasets, and a new sub-dataset is used for each fine-tuning iteration. Before each fine-tuning, we call the latest current model to generate a unified preference dataset for a portion of the samples. This is to ensure that the samples conform to the model's preferences. At the same time, we use a teacher VLM to compare the current samples with the samples generated by the model itself, taking the best one as the training sample. This is to maintain the high quality of the data and keep the sample noise within a tolerable range for the model.

\section{More Experimental Details}
\subsection{Base Model.} This work utilizes Qwen2.5-VL as the base model, which unifies image/video understanding and reasoning tasks, demonstrating the mutual benefits of multi-task learning, and achieves excellent performance on various visual benchmarks. However, its reasoning is limited to direct responses or shallow thinking, lacking the long-term and structured self-verification and self-rectification capabilities of Chain-of-Thought (CoT) reasoning. This may lead to reduced accuracy and weakened interpretability in complex scenarios. Inspired by this work, we further activate the potential long-term self-verification and self-rectification reasoning capabilities of Qwen2.5-VL in various visual reward tasks, aiming to improve the accuracy and robustness of reward signals.

\subsection{Model Baselines.}
We compared our method against a series of strong model baselines in the image domain, covering understanding and reasoning tasks.

\noindent\textbf{GPT-4V} is a powerful multimodal model from OpenAI, renowned for its excellent image understanding and visual question answering (VQA) capabilities. It can deeply analyze image content, provide detailed descriptions, recognize objects and scenes, and perform complex reasoning based on visual information.

\noindent\textbf{GPT-4o-20240513} ~\cite{hurst2024gpt} is OpenAI's latest flagship model, marking a step towards more natural and efficient human-computer interaction. It natively integrates text, audio, and visual processing capabilities, understanding and generating combinations of these modalities at extremely high speed, enabling smoother, more expressive real-time conversations and multimodal outputs.

\noindent\textbf{Claude-3.5-Sonnet} is Anthropic's latest and fastest model, excelling in visual reasoning, code generation, and text editing. It is particularly adept at tasks requiring complex instructions and visual analysis, such as generating code from sketches or interpreting charts. It maintains high performance while offering better cost-effectiveness than its predecessor flagship model.

\noindent\textbf{Gemini-1.5-Pro} is notable for its breakthrough million-token long context window, enabling it to process and understand massive amounts of text, code, audio, and even hours-long video content. It excels in multimodal understanding and long-context reasoning, capable of extracting key information from vast data and performing in-depth analysis.

\noindent\textbf{InternVL2.5} ~\cite{chen2025expandingperformanceboundariesopensource} , from the Shanghai AI Laboratory, is a powerful open-source multimodal large model. It has achieved leading results on multiple international authoritative multimodal and visual benchmarks, particularly excelling in tasks like fine-grained image recognition, OCR, and visual question answering (VQA), demonstrating its profound strength in understanding complex visual information.

\noindent\textbf{MiniCPM-V2.6} ~\cite{hu2024minicpmunveilingpotentialsmall} , jointly developed by Tsinghua University and ModelBest, is a lightweight yet high-performance open-source multimodal model. It stands out for its small size ("on-device large model") and efficient performance, excelling in general visual understanding, precise OCR, and low-hallucination dialogue, making it well-suited for deployment on resource-constrained devices.

\noindent\textbf{InternVL3} ~\cite{zhu2025internvl3exploringadvancedtraining} is the latest upgraded version in the InternVL series, further increasing its parameter scale and performance ceiling. It has reached new heights in processing ultra-high-resolution images, performing deeper visual reasoning, and nuanced scene understanding, continuously breaking records in top-tier multimodal benchmarks and representing the cutting edge of open-source multimodal research.

\noindent\textbf{Qwen2-VL} ~\cite{wang2024qwen2vlenhancingvisionlanguagemodels}, part of Alibaba's Tongyi Qianwen series, is a multimodal model with strong general visual understanding capabilities and excellent Chinese/English support. It can process image inputs to perform tasks like VQA and image description, and has demonstrated competitive performance on various benchmarks, making it suitable for a wide range of practical application scenarios.

\noindent\textbf{Cambrian} ~\cite{tong2024cambrian1fullyopenvisioncentric} represents an exploratory research initiative focused on vision-centric multimodal large language models, with full openness as a core tenet. Its key characteristic is its commitment to providing a transparent platform for the research community to deeply investigate how visual information can be more effectively integrated with language models. It emphasizes the deep integration and optimization of visual components, rather than simple concatenation, and encourages experimentation with diverse visual encoders, connection modules, and training strategies to drive the advancement of vision-centric multimodal models.

\noindent\textbf{LLaVA-OneVision} ~\cite{li2024llavaonevisioneasyvisualtask} is built upon the LLaVA architecture and achieves this by employing a powerful and highly generalizable unified visual encoder (the "OneVision" component). This enables the model to more easily transfer its learned visual knowledge and skills to new, unseen visual tasks without requiring extensive task-specific fine-tuning, underscoring its robustness and generality in visual representation learning for more efficient cross-task understanding.

\noindent\textbf{Qwen2.5VL} ~\cite{bai2025qwen25vltechnicalreport} is an upgraded version of Qwen2-VL, featuring significant enhancements in both the depth and breadth of visual understanding. It has specifically improved capabilities in fine-grained recognition, complex scene reasoning, and multi-turn visual dialogue, allowing it to more accurately capture and understand subtle differences and complex relationships in images, while maintaining excellent generality and multilingual support.

\noindent\textbf{Our SVSR'Qwen2.5-VL } extends UnifiedReward by integrating explicit long CoT reasoning across both visual understanding and generation tasks. Through a three-stage training pipeline—including cold start to learning CoT reward format, rejection sampling for unified CoT reward generalization fine-tuning, and GRPO for unified CoT reward reinforcement fine-tuning, the model achieves stronger accuracy and interpretability in reward assessment. It also generalizes well without explicit reasoning, leveraging implicit CoT capabilities for robust performance.

\subsection{Evaluation Benchmarks}

\noindent\textbf{MathVista} ~\cite{LuMathVista2023} serves as a diverse benchmark for evaluating image understanding capabilities, featuring 1,250 carefully curated samples across general vision-language queries, hallucination detection, and complex reasoning. To ensure robust evaluation, response orders are randomly shuffled during testing.

\noindent\textbf{DynaMath} ~\cite{DynaMathDataset} provides large-scale video-caption pairs and human preference data, covering various aspects of video understanding such as temporal reasoning, spatial relations, and factual grounding. We 3K for evaluation in our reward modeling experiments.

\noindent\textbf{MathVerse} ~\cite{MathVerseDataset} is a multimodal generation benchmark designed to assess how well models align with human preferences across image and video generation tasks. We adopt its image and video generation subsets for evaluating generative reward performance.

\noindent\textbf{MMstar} ~\cite{MMStarDataset} offers a large-scale benchmark tailored for evaluating video reward models, consisting of 26.5k video pairs labeled by humans. Each pair is ranked according to multiple criteria, and we use the Overall Quality scores to benchmark the performance of the model.

\noindent\textbf{MMvet} ~\cite{MMVetDataset} offers a large-scale benchmark tailored for evaluating video reward models, consisting of 26.5k video pairs labeled by humans. Each pair is ranked according to multiple criteria, and we use the Overall Quality scores to benchmark the performance of the model.

\noindent\textbf{AI2D} ~\cite{AI2DDataset2016} offers a large-scale benchmark tailored for evaluating video reward models, consisting of 26.5k video pairs labeled by humans. Each pair is ranked according to multiple criteria, and we use the Overall Quality scores to benchmark the performance of the model.

\section{More Qualitative Cases}
\begin{figure*}[t!]
    \centering
    \includegraphics[width=\textwidth]{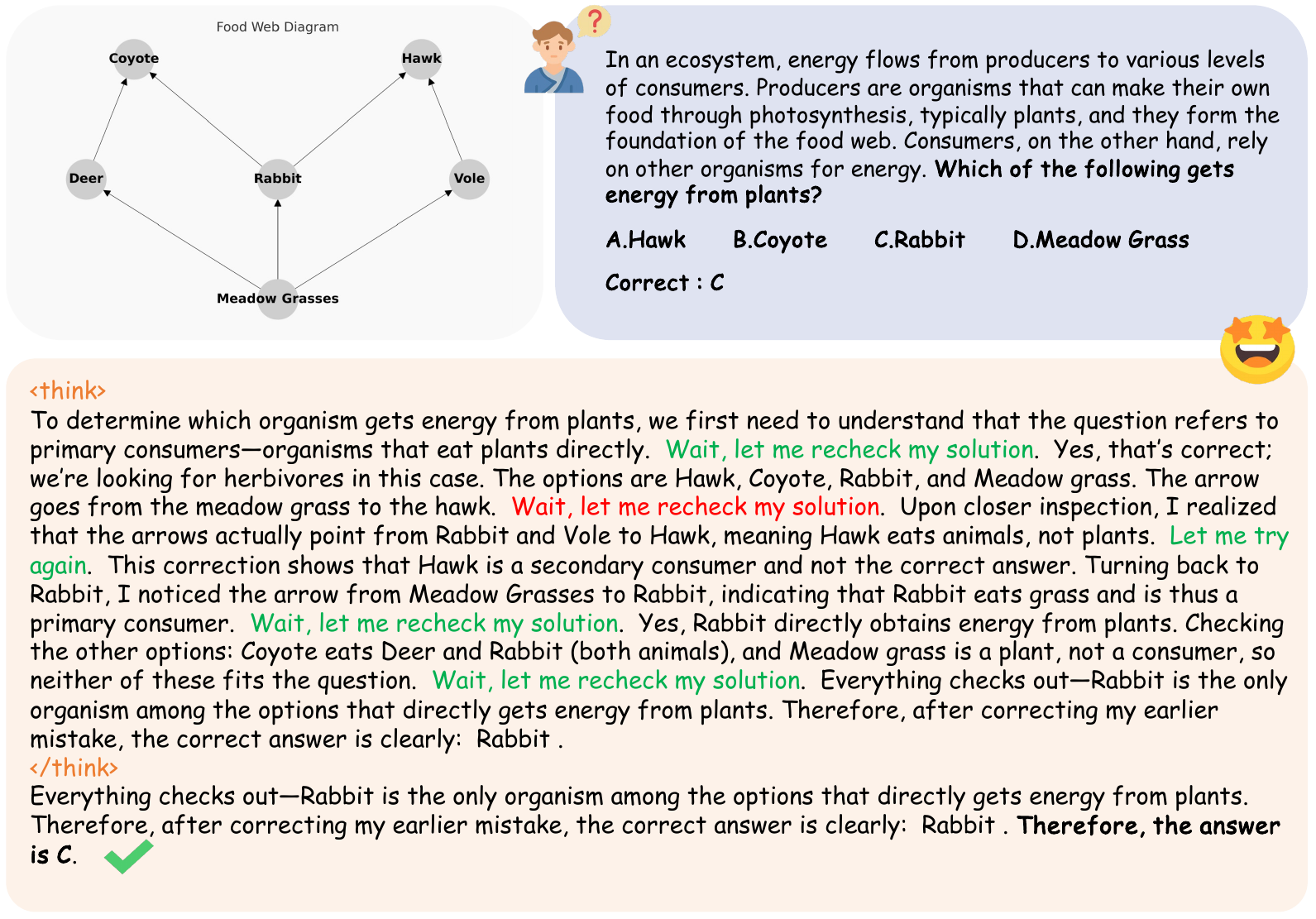}
    \caption{This figure illustrates the model's reasoning process when applying the SVSR framework to a science question about energy flow in ecosystems. The example demonstrates the three key phases of our approach: initial reasoning, where the model first analyzes which organism gets energy from plants; self-verification, where it explicitly checks its solution (indicated by "Wait, let me recheck my solution"); and self-rectification, where it identifies and corrects its misconception about the direction of energy flow in the food web. The model ultimately arrives at the correct answer (C. Rabbit) after successfully implementing the verification-correction cycle, showcasing how SVSR enables structured, multi-step reasoning with built-in error detection and correction.}
    \label{fig:appendix}
\end{figure*}

We provide more qualitative cases for vision tasks in the Figure~\ref{fig:appendix}.

\section{Limitations}
\label{limitations}
While our approach introduces self-verification and self-rectification to improve model reasoning capabilities, this inevitably increases the reasoning time during the inference process. However, we show that once the model has mastered self-verification and self-rectification reasoning, it can leverage implicit reasoning to improve the accuracy of answers even without explicitly generating self-verification and self-rectification trajectories. This indicates a strong internalization of the reasoning process. 

\section{Future Works}
In future work, we plan to explore shorter or more efficient self-verification and self-rectification reasoning formats to further optimize efficiency without compromising reasoning quality. Furthermore, although our reinforcement fine-tuning successfully leveraged a small amount of high-quality data to activate the model's latent self-verification and self-rectification reasoning abilities, previous research has shown that reinforcement learning cannot fundamentally expand the model's capabilities: it can only amplify the potential already acquired during supervised fine-tuning (SFT). Therefore, to further advance the boundaries of self-verification and self-rectification, scaling up high-quality self-verification and self-rectification formatted data remains a promising direction.

\section{Societal Impacts}
\label{impact}
Our work introduces a unified multimodal training framework capable of high-quality, interpretable reasoning across various multimodal tasks. This advancement can significantly improve the alignment of multimodal models with human preferences in practical applications such as AI-assisted content creation and education. By enhancing the accuracy and interpretability of model reasoning, our approach contributes to more transparent and controllable AI behavior, potentially increasing public trust in multimodal technologies. However, as multimodal models become more powerful and versatile, they also carry the risk of misuse, potentially reinforcing harmful biases through model-generated content, especially if the training data or preference annotations reflect subjective or biased human values. We encourage future work to further investigate the ethical implications of large-scale self-verification and self-rectification, and to incorporate fairness-aware training strategies.

\section{Ethical Statement}
\label{ethical}
In this work, we declare our commitment to ethical research practices and responsible innovation. To the best of our knowledge, this research does not involve any data, methods, or applications that would raise ethical concerns. All experiments and analyses were conducted in accordance with established ethical guidelines, ensuring the integrity and transparency of our research process.

\end{document}